\documentclass[10pt,twocolumn,letterpaper]{article}
\usepackage[pagenumbers]{cvpr}
\usepackage{booktabs,array,xcolor}
\usepackage{graphicx}
\usepackage{cuted}

\definecolor{saffron}{rgb}{0.95,0.75,0.2}

\usepackage{xcolor}
\usepackage{circledsteps}

\definecolor{circnumcolor}{HTML}{39485F}

\newcommand{\circnumours}[1]{\Circled[fill color=circnumcolor,inner color=white,outer color=circnumcolor]{\scriptsize #1}}

\definecolor{colorC}{HTML}{00CCCC}

\usepackage{tcolorbox}
\tcbuselibrary{breakable}

\definecolor{cvprblue}{rgb}{0.21,0.49,0.74}
\definecolor{linkpurple}{RGB}{180,0,180}
\usepackage[pagebackref,breaklinks,colorlinks,allcolors=cvprblue]{hyperref}

\hypersetup{
  pdftitle={Video Models are Excellent Motion Models},
  pdfauthor={Kian Izadpanah, Ali Mahdavi Amiri, Daniel Cohen-Or,
             Yiorgos Chrysanthou, Andreas Aristidou},
  pdfsubject={3D Human Motion Generation}
}

\title{Video Models are Excellent Motion Models}

\author{
Kian Izadpanah$^{1,2}$ \qquad
Ali Mahdavi Amiri$^{3}$ \qquad
Daniel Cohen-Or$^{4}$ \\
Yiorgos Chrysanthou$^{1,2}$ \qquad
Andreas Aristidou$^{1,2}$ \\[8pt]
$^{1}$University of Cyprus \qquad
$^{2}$CYENS Centre of Excellence \\
$^{3}$Simon Fraser University \qquad
$^{4}$Tel Aviv University
}

\begin{document}
\maketitle
\setlength{\stripsep}{0pt}
\begin{strip}
 \centering
 \includegraphics[width=.76\textwidth]{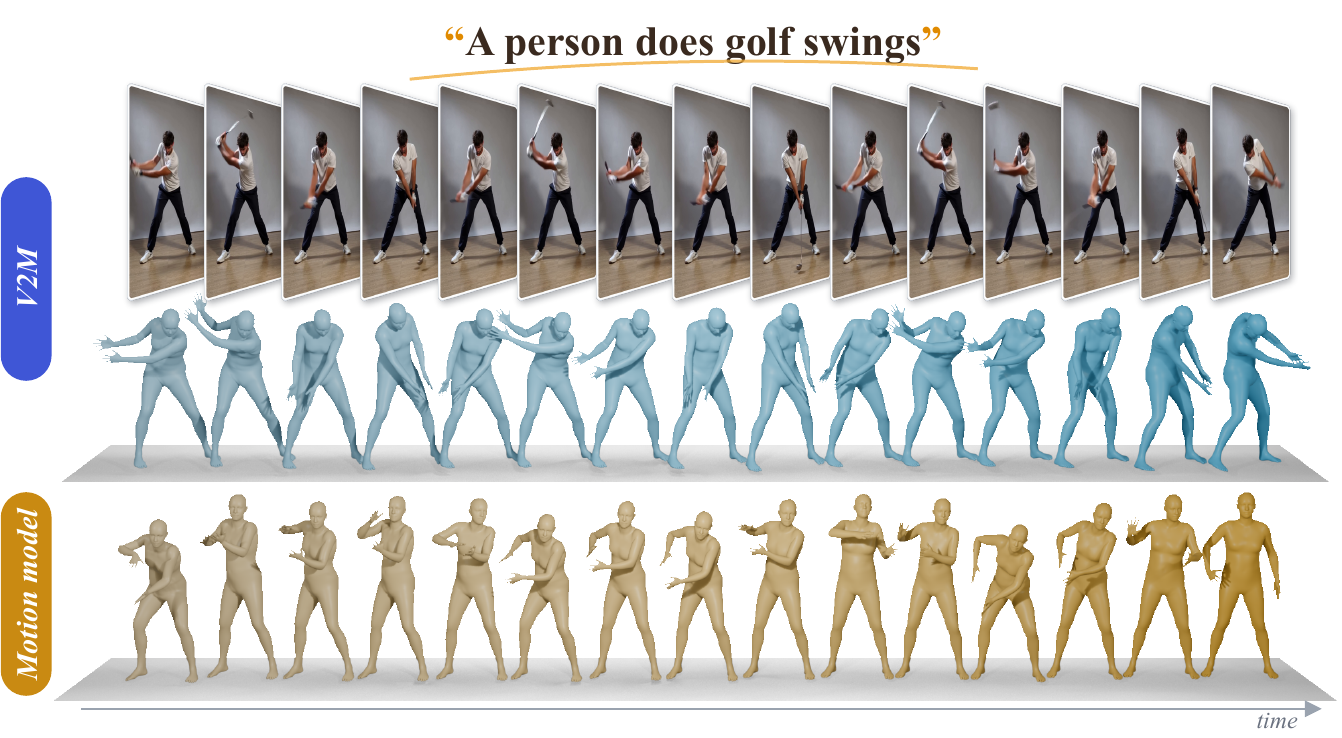}
 \captionsetup{hypcap=false}
 \captionof{figure}{Given the same text prompt, V2M generates a more complete and coordinated golf swing than a dedicated motion model, illustrating the potential of video foundation models to produce richer and more semantically coherent 3D human motion.}
 \captionsetup{hypcap=true}
 \label{fig:teaser}%
 \vspace{10pt}
\end{strip}
 
\begin{abstract}
Video foundation models have emerged as powerful generators of dynamic visual content, exhibiting remarkable capabilities in synthesizing complex actions and interactions. We argue that these models have also become \textbf{excellent motion generators}, with motion priors that are richer and more general than those learned by dedicated motion models. Their advantage extends beyond visual realism and is manifested in two important aspects: \textbf{motion naturalness}, including coherent body coordination, physical plausibility, and realistic interactions with objects and the environment; and \textbf{motion scope}, encompassing a much broader range of activities and semantic coherence, particularly uncommon, complex, and compound actions. To investigate this position, we introduce a common evaluation framework that compares video and dedicated motion models through three complementary settings: text-only motion generation, reconstruction of video-generated motion from sparse keyframes, and video-based enrichment of motion-model outputs. By converting all outputs into a shared 3D skeletal representation, we evaluate motion naturalness, semantic alignment, action coverage, physical motion characteristics, and fidelity to motion constraints. Our results suggest that video foundation models can generate more natural and semantically meaningful movements, particularly for complex and compound actions. Moreover, we investigate motion-conditioned generation through loose skeletal control, finding that dedicated motion models do not consistently improve video-generated motion, while video models can enrich skeletal animations using sparse motion guidance. Our findings suggest a shift in the role of specialized motion models, \textbf{from generating complete animations to guiding and controlling the richer motion capabilities already embedded in video foundation models.} 

\textbf{Project Page:} 
\href{https://video-beats-motion.github.io/}
{\textcolor{linkpurple}{video-beats-motion.github.io}}
\end{abstract}    
\section{Introduction}
\label{sec:introduction}

Recent advances in video generation have brought us closer to the vision of generative world models that capture not only how the world looks, but also how it behaves and evolves over time~\cite{brooks2024video}. Trained on large and diverse video datasets, modern video foundation models have demonstrated remarkable capabilities in generating complex human activities, coordinated movements, and interactions between people, objects and the environment. These capabilities suggest that their learned representations extend beyond visual appearance, capturing useful knowledge about human motion and the relationships that govern human behavior. This raises an interesting question: \emph{can the motion knowledge embedded in video foundation models be used to generate more natural and semantically meaningful 3D human motion?} 

Dedicated generative motion models have also made significant progress in synthesizing realistic and controllable human movements from text descriptions, sparse keyframes, and other control signals~\cite{tevet2023human,cohan2024flexible,rempe2026kimodo}. By explicitly modeling skeletal poses and trajectories, they provide precise control over character animation. An important difference between motion and video models, however, lies in the scale and diversity of their training data.

Although motion models can benefit from larger training datasets, collecting diverse, high-quality, and semantically annotated 3D motion data remains challenging and expensive. In contrast, video data are substantially more abundant, allowing video foundation models to be trained on tens of thousands of hours of recordings covering a broader range of human activities, objects, and environments~\cite{opensora}. Although video and skeletal motion data are not directly comparable, the greater availability and diversity of video data may provide richer motion priors, particularly for uncommon and complex activities, human-object interactions, and compound actions that are less represented in structured motion datasets. 

In this work, we argue that \emph{video foundation models have become stronger and more general motion generators than dedicated motion models}, both in motion naturalness and the breadth of actions they can represent. First, video models have learned rich motion priors that capture not only human movement but also relationships between body parts, objects, and the environment. This knowledge enables more coherent and physically plausible movements, such as maintaining a consistent grasp when carrying a box, coordinating hand and foot placement when climbing a ladder, or combining actions, such as walking while drinking coffee and talking on a phone. Second, training on substantially larger and more diverse datasets expose video models to a broader range of activities and interactions, including those less represented in structured motion datasets. Importantly, our claim concerns the quality and semantic consistency of the underlying 3D motion, rather than the visual realism of generated videos. As illustrated in \cref{fig:teaser}, video foundation models can generate richer, more natural, and semantically coherent 3D human motion than dedicated motion models given the same text prompt.

To investigate this position, we introduce a common evaluation framework that converts video- and motion-model outputs into a shared 3D skeletal representation. We examine three complementary experimental settings. First, \emph{text-only generation} compares the ability of both model families to independently generate natural and semantically meaningful movements from the same prompt. Second, \emph{video-to-motion} examines whether dedicated motion models can reconstruct or improve video-generated motion conditioned on a small set of keyframes. Third, \emph{motion-to-video} investigates whether video models can enrich motion-model outputs under sparse skeletal guidance while retaining their intended motion structure. Together, these settings allow us to examine motion range, naturalness, and the complementary roles of video and motion models.

Our experiments provide evidence that video foundation models generate more natural and semantically meaningful movements, particularly for complex and compound actions. While dedicated motion models do not consistently improve video-generated motion through sparse-keyframe reconstruction, video models can enrich skeletal animations through loose control while preserving their original intent. We support these findings through semantic, perceptual, and quantitative motion evaluations. Our contributions offer a new perspective on video foundation models as motion generators, a unified evaluation framework, and empirical evidence of their richer motion priors. Our findings suggest a broader shift in motion generation, where video foundation models increasingly contribute rich semantic understanding and motion priors for generating natural and expressive animations, while dedicated motion models offer complementary capabilities for refining and controlling the generated motion.

\section{Related Work}

\subsection{Human Motion Generation}

Recent advances in generative models have significantly improved 3D human motion synthesis from natural language. Early approaches, e.g., TEMOS~\cite{petrovich2022temos}, MDM~\cite{tevet2023human}, and MotionDiffuse~\cite{zhang2024motiondiffuse}, established frameworks for text-to-motion generation, while T2M-GPT~\cite{zhang2023t2mgpt}, MotionGPT~\cite{jiang2023motiongpt}, and MoMask~\cite{guo2024momask} explored discrete motion representations and language-based generation. More recent methods, including Motion Mamba~\cite{zhang2024motionmamba}, BAMM~\cite{pinyoanuntapong2024bamm}, MotionLCM~\cite{dai2024motionlcm}, and EMDM~\cite{zhou2024emdm}, have improved motion quality, efficiency, and controllability. K-Mask~\cite{ahmed2026kmask} further supports fine-grained body-part control, while CondMDI~\cite{cohan2024flexible} enables flexible motion in-betweening from sparse keyframes. Kimodo~\cite{rempe2026kimodo} extends controllable motion generation through larger-scale training. These advances provide a strong foundation for comparing dedicated motion models with video foundation models.

Despite these advances, generating complex and compound actions remains challenging, particularly when multiple movements must be performed simultaneously. Models may prioritize a dominant action while neglecting other components of the prompt, resulting in incomplete or semantically inconsistent motion~\cite{athanasiou2023sinc,sala2026multiact,jiang2026motionadapter}. Similar challenges arise with sequential actions, where models may omit movements, violate their order, or produce unnatural transitions~\cite{athanasiou2022teach,barquero2024seamless,hong2026eventt2m}, while others struggle to preserve global semantics~\cite{zheng2026moscale}. Further challenges include satisfying spatial and physical constraints, particularly in human-object interactions, leading to foot sliding, unnatural coordination, or ground penetration. Overall, generating diverse, natural, and semantically faithful complex and compound movements remains an open challenge.

A contributing factor is the limited availability of diverse, high-quality, and semantically annotated 3D motion data. Widely used datasets such as HumanML3D~\cite{guo2022generating} contain approximately 29 hours of motion, while Motion-X~\cite{lin2023motionx} expanded this to approximately 144 hours. Recent efforts have further increased this scale, including Kimodo, trained on approximately 700 hours of motion capture~\cite{rempe2026kimodo}, and MotionMillion, containing more than 2,000 hours~\cite{fan2025gotozero}. VideoMDM~\cite{mann2026videomdm} has also explored learning 3D motion priors from 2D poses extracted from videos. Nevertheless, uncommon activities, complex interactions, and compound actions remain less extensively represented in structured motion datasets. Moreover, many annotations describe only the dominant action, providing limited information about simultaneous movements, temporal relationships, and interaction constraints. Thus, the challenge concerns not only data scale, but also motion diversity and semantic richness.

\subsection{Video Foundation Models}

Video foundation models have advanced through large-scale generative training, demonstrating remarkable capabilities in synthesizing realistic scenes, human activities, and complex interactions. Models such as Sora~\cite{brooks2024video}, Movie Gen~\cite{polyak2024moviegen}, CogVideoX~\cite{yang2024cogvideox}, HunyuanVideo~\cite{kong2024hunyuanvideo}, Open-Sora~\cite{opensora}, and Wan~\cite{wang2025wan} have substantially improved video quality, temporal consistency, and text alignment. Recent developments, including MiniMax-H3~\cite{minimax2026h3} and LTX-2~\cite{hacohen2026ltx2}, have further advanced generation, complex instruction following, and flexible conditioning. Of particular relevance to our work, MiniMax-H3 and LTX-2.5 support motion-guided video generation through ControlNet Union 2.0~\cite{alibabapai2026controlnetunion} and IC-LoRA Union Control~\cite{benyosef2026avcontrol} conditioning, respectively. These mechanisms enable existing animations to provide structural guidance while video foundation models generate additional motion details and interactions.

A key difference between video and motion models lies in the scale and diversity of training data. Video models learn from visual observations of human movements, objects, interactions, and environments, potentially capturing richer contextual motion priors. For instance, Open-Sora reports training on approximately 40,000 hours of video~\cite{opensora}. Although these datasets are not directly comparable, their differences motivate investigating whether video models generate more natural and semantically meaningful 3D motion, particularly for complex and compound actions.

Recent studies have explored video models as sources of motion priors for 3D animation. Animating the Uncaptured~\cite{benedi2026animating} transfers motion from synthesized videos to 3D character meshes, while CoMoVi~\cite{zhao2026comovi} and EchoMotion~\cite{yang2026echomotion} explore joint video and 3D motion generation. SViMo~\cite{dang2025svimo} combines visual and motion priors for human-object interactions, while Ani3DHuman~\cite{sun2026ani3dhuman} exploits video diffusion for non-rigid character animation. Recent approaches increasingly integrate video and 3D motion generation. Rotstein~et~al.~\cite{rotstein2026jca} and Rahamim~et~al.~\cite{rahamim2026all} explore joint video-motion synthesis, while ISA4D~\cite{shao2025isa4d} enables controllable human video generation, and R-DMesh~\cite{wu2026rdmesh} and ActionMesh~\cite{sabathier2026actionmesh} extend video-guided animation to 3D meshes. These approaches demonstrate the growing potential of video models for motion synthesis.

Advances in human pose estimation and 3D motion reconstruction have also made it increasingly feasible to recover skeletal animations from videos. Methods such as DWPose~\cite{yang2023effective} provide detailed whole-body keypoints, while WHAM~\cite{shin2024wham} and GVHMR~\cite{shen2026gvhmr} recover 3D human motion and global trajectories from monocular videos. More recent approaches, including HumanMM~\cite{zhang2025humanmm}, PEAR~\cite{wu2026pear}, and OnlineHMR~\cite{zhao2026onlinehmr}, have further advanced motion reconstruction. Despite remaining challenges involving occlusions, rapid movements, and complex interactions, these developments provide an increasingly practical connection between video generation and editable 3D animation. This motivates our investigation of video foundation models as general-purpose motion generators.

\begin{figure*}[h]
\centering
\includegraphics[width=\textwidth]{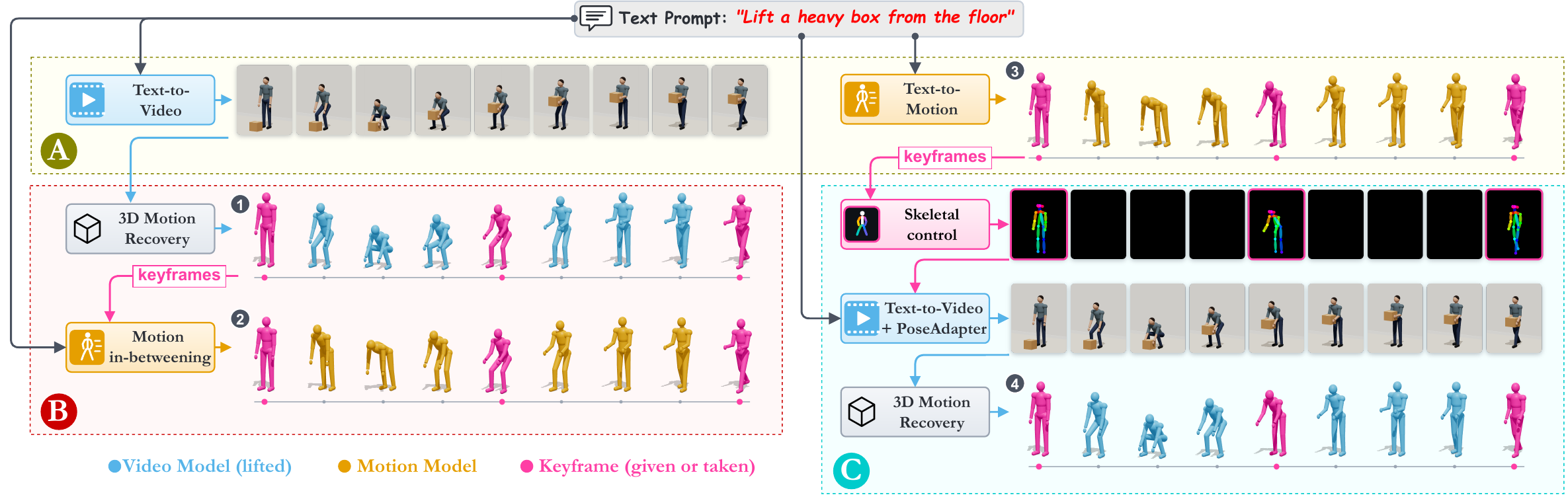}
\caption{\textbf{Comparison framework.} A common text prompt is used across video and dedicated motion models, and all outputs are converted into the same 3D skeletal representation. \emph{Left:} The video model generates a video, from which 3D motion is recovered through lifting~\circnumours{1}. A dedicated motion model then reconstructs the intermediate movements from $K$ selected keyframes~\circnumours{2}. \emph{Right:} The dedicated motion model generates a skeletal animation~\circnumours{3}, whose $K$ keyframes provide loose skeletal guidance to the video model. The resulting video is reconstructed into 3D motion~\circnumours{4}. These outputs define three complementary comparisons: \textcolor{olive}{\textbf{A}} (\circnumours{1} vs.\ \circnumours{3}) evaluates independent text-only generation; \textcolor{red}{\textbf{B}} (\circnumours{1} vs.\ \circnumours{2}) examines whether dedicated motion models can improve video-generated motion; and \textcolor{colorC}{\textbf{C}} (\circnumours{3} vs.\ \circnumours{4}) investigates whether video models can enrich skeletal animations while respecting sparse motion guidance. Blue and orange indicate video- and motion-model outputs, respectively, while pink marks the selected keyframes.}
\label{fig:pipeline}
\vspace{-7pt}
\end{figure*}

\section{Comparison Framework}
\label{sec:framework}

Our central position is that \emph{video foundation models have become stronger and more general motion generators than dedicated motion models}, benefiting from richer motion priors. However, comparing these models is challenging because motion models generate skeletal animations, while video models produce image sequences requiring 3D motion reconstruction. We therefore introduce a common evaluation framework (\cref{fig:pipeline}) that converts all outputs into a shared 3D skeletal representation. We investigate three complementary settings: (A) independent text-only generation, (B) reconstruction of video-generated motion using motion models conditioned on sparse keyframes, and (C) enrichment of motion-model outputs through video generation under loose skeletal control. These comparisons assess motion range, naturalness, semantic alignment, action coverage, and the complementary roles of both model families.

\subsection{Research Hypotheses}
\label{sec:hypotheses}
We formulate three hypotheses to assess our claim.

\vspace{2pt}
\noindent{\textbf{H1 -- Motion Range and Semantic Alignment.}} Compared with dedicated motion models, video foundation models can generate a broader range of semantically meaningful human activities, particularly uncommon and complex movements, simultaneous or sequential actions, and interactions with objects or the environment. We primarily investigate this hypothesis in Experiment A, evaluating semantic alignment and the coverage of individual actions specified in the prompt.

\vspace{2pt}
\noindent{\textbf{H2 -- Motion Naturalness and Coherence.}} Video foundation models can generate more natural and contextually appropriate movements than dedicated motion models, with better body coordination, temporal coherence, and consistency with the intended action. We investigate this hypothesis through perceptual comparisons and quantitative motion-quality metrics across all three experiments.

\vspace{2pt}
\noindent{\textbf{H3 -- Motion Enrichment under Sparse Guidance.}} The rich motion priors of video foundation models can be exploited through loose skeletal control to enrich existing animations with more natural and semantically meaningful movements while retaining the structural guidance provided by sparse keyframes. We primarily investigate this hypothesis in Experiment C, evaluating motion enrichment through perceptual comparisons and consistency with the supplied guidance through keyframe fidelity. Experiment B provides a complementary assessment of whether dedicated motion models can improve video-generated motion through sparse-keyframe reconstruction.

\subsection{Common Representation and Reconstruction}
\label{sec:representation}
We represent motion as a sequence $\mathbf{x}=(\mathbf{p}_t)_{t=1}^{T}$, where each pose $\mathbf{p}_t\in\mathbb{R}^{J\times3}$ contains the positions of $J=22$ SMPL body joints~\cite{SMPL:2015} at frame $t$. All sequences are converted to a gravity-aligned, $Y$-up coordinate system at 30\,fps, preserving their original duration. We normalize the initial root position and use consistent skeletal proportions. 

Let $\mathcal{V}$ denote a video foundation model, $\mathcal{M}$ a dedicated motion model, and $\mathcal{L}$ the video-to-motion reconstruction operator. Given a text prompt $c$, the generated motions are:
\begin{equation}
    \mathbf{x}^{V}=\mathcal{L}(\mathcal{V}(c)), \qquad
    \mathbf{x}^{M}=\mathcal{M}(c).
    \label{eq:motionoutputs}
\end{equation}

We use MiniMax-H3~\cite{minimax2026h3} for text-only video generation and its ControlNet Union 2.0~\cite{alibabapai2026controlnetunion} conditioning mechanism for skeletal-guided video generation. 
We recover 3D motion using GVHMR~\cite{shen2026gvhmr}, while DWPose~\cite{yang2023effective} provides skeletal guidance when needed. Our dedicated motion models include Kimodo~\cite{rempe2026kimodo} and MotionGPT3~\cite{motiongpt3} for text-to-motion generation, and Kimodo and CondMDI~\cite{cohan2024flexible} for motion in-betweening. All outputs are converted to the same skeletal representation before evaluation.

\subsection{Comparison Pipeline}
\label{sec:pipeline}

As illustrated in~\cref{fig:pipeline}, the three experiments examine complementary aspects of our position. Exp. A directly compares independent motion generation, while Exp. B and C use sparse keyframes to guide motion generation and enable comparisons of motion quality relative to the same guidance within each experiment.

For B and C, we select $K=5$ uniformly spaced full-body keyframes, including the first and last frames of each sequence. Let $\mathbf{x}_{\mathcal{K}}$ denote these keyframes of motion $\mathbf{x}$, and let $\mathcal{R}_{\mathrm{kf}}(\mathbf{x}_{\mathcal{K}})$ denote their projection into 2D skeletal guidance for video generation. Our comparisons are defined as:
\begin{equation}
    \begin{aligned}
        \text{A:}\quad&
        \mathbf{x}^{V} \quad \text{vs.} \quad \mathbf{x}^{M},\\
        \text{B:}\quad& \mathbf{x}^{V} \quad \text{vs.}\quad \mathcal{M}(c,\mathbf{x}^{V}_{K}),\\
        \text{C:}\quad& \mathbf{x}^{M} \quad \text{vs.}\quad \mathcal{L}\big(\mathcal{V}(c,\mathcal{R}_{\mathrm{kf}}(\mathbf{x}^{M}_{K})) \big).
    \end{aligned}
    \label{eq:settings}
\end{equation}

\vspace{2pt}
\noindent{\textbf{A. Text-only Generation (H1, H2).}} Both models independently generate motion from the same text prompt, without skeletal guidance, allowing direct comparisons of naturalness, semantic alignment, and action complexity. This setting directly investigates whether video foundation models possess broader and richer motion-generation capabilities.

\vspace{2pt}
\noindent{\textbf{B. Video-to-Motion (H2, H3).}} We extract five keyframes from the reconstructed video motion to condition dedicated motion models, alongside the original text prompt, examining whether the resulting animations preserve or improve motion naturalness and semantic consistency.

\vspace{2pt}
\noindent{\textbf{C. Motion-to-Video (H2, H3).}} Five keyframes from a motion-model animation provide sparse, loose 2D guidance for video generation, alongside the original text prompt. We examine whether the resulting motion becomes more natural and semantically meaningful while preserving the intended structure conveyed by the keyframes.

\subsection{Sparse-Keyframing and Loose Control}
\label{sec:loosecontrol}

Experiments B and C investigate conditional motion generation through sparse keyframes, but differ in how these constraints are applied. In B, dedicated motion models receive 3D full-body keyframes to generate intermediate skeletal movements. In C, the selected poses are projected into 2D and provided to the video model through ControlNet Union 2.0, without additional training. 

We refer to the latter as \emph{loose skeletal control}: selected keyframes provide structural and temporal guidance without prescribing the complete animation. This shared guidance helps constrain the intended action, enabling more meaningful comparisons of motion quality while allowing the video model to generate intermediate movements. Our objective is to exploit its rich motion priors to enrich animations with more natural and semantically meaningful motion. Keyframe fidelity measures consistency with the guidance, rather than an absolute measure of motion quality.

Importantly, B and C use different conditioning representations (3D versus 2D) and are therefore complementary rather than directly equivalent comparisons. Furthermore, since ControlNet Union 2.0 was developed using dense skeletal guidance, unspecified intermediate poses are not necessarily unconstrained. These differences, together with potential 3D reconstruction errors, are considered when interpreting the results.

\subsection{Evaluation Criteria and Metrics}
\label{sec:evaluation}

We evaluate the underlying 3D motion rather than the visual appearance of generated videos, combining VLM-based comparisons, human perceptual evaluation, and quantitative motion analysis. All animations are represented using a common skeletal structure and consistent rendering conditions. Detailed metric definitions and procedures are provided in \cref{sec:supp_metrics}.

\vspace{2pt}
\noindent{\textbf{Semantic Alignment and Action Coverage (H1, H3).}} We use Qwen3-VL-32B-Instruct~\cite{Qwen3-VL} as an automated judge to compare anonymized motion pairs in terms of semantic alignment with the text prompt and perceived naturalness. We report win/tie/loss (W/T/L) percentages, using reversed presentation orders to reduce bias. Our perceptual study additionally evaluates action coverage, examining whether the generated movements perform the requested actions, particularly for complex and compound prompts.

\vspace{2pt}
\noindent{\textbf{Motion Naturalness and Coherence (H2).}} We assess perceived naturalness through VLM-based comparisons and human evaluations. Quantitative measurements include positional jerk, angular acceleration, estimated ground-contact coverage, and foot sliding. These metrics characterize motion dynamics and potential artifacts.

\vspace{2pt}
\noindent{\textbf{Motion Fidelity (H3).}} In Exp. B and C, we assess consistency of the supplied skeletal guidance using Mean Per Joint Position Error (MPJPE), Procrustes-Aligned MPJPE (PA-MPJPE), and geodesic joint-rotation errors, including root orientation. These metrics quantify consistency with the selected keyframes rather than motion quality itself, since loose control permits deviations that may improve naturalness and semantic coherence.

\vspace{2pt}
\noindent{\textbf{Reconstruction Considerations (H1--H3).}} Since video-generated motion requires 3D reconstruction, measured artifacts may originate from reconstruction rather than generation. We therefore interpret quantitative results alongside VLM and human evaluations, considering possible reconstruction errors.

\vspace{2pt}
\noindent{\textbf{Human Perceptual Evaluation (H1--H3).}} We conduct a human perceptual study assessing motion naturalness, semantic alignment, action coverage, and visible motion artifacts. These judgments complement automated semantic evaluation and quantitative motion-quality metrics, particularly for complex and compound actions.

\section{Evaluation Results}
\label{sec:experiments}

\begin{table}[t]
\centering
\caption{\textbf{Pairwise VLM evaluation.} W/T/L (\%) for V2M against each baseline, grouped by common and complex actions.}
\label{tab:pairwise_results}
\footnotesize
\setlength{\tabcolsep}{2pt}
\renewcommand{\arraystretch}{1.08}
\begin{tabular}{@{}lllcc@{}}
\toprule
\textbf{Exp.} & \textbf{vs.} & \textbf{Group} & \textbf{Text W/T/L} & \textbf{Natural. W/T/L} \\
\midrule
A & Kimodo & Common & 57.6\,/\,36.4\,/\,6.10 & 57.6\,/\,24.2\,/\,18.2 \\
  & & Complex & 59.4\,/\,21.9\,/\,18.8 & 56.2\,/\,25.0\,/\,18.8 \\
  & M-GPT3 & Common & 39.4\,/\,33.3\,/\,27.3 & 36.4\,/\,33.3\,/\,30.3 \\
  & & Complex & 53.1\,/\,25.0\,/\,21.9 & 53.1\,/\,21.9\,/\,25.0 \\
\midrule
B & Kimodo & Common & 12.5\,/\,81.2\,/\,6.20 & 12.5\,/\,81.2\,/\,6.20 \\
  & & Complex & 12.5\,/\,84.4\,/\,3.10 & 9.4\,/\,87.5\,/\,3.10 \\
  & CondMDI & Common & 18.8\,/\,68.8\,/\,12.5 & 15.6\,/\,62.5\,/\,21.9 \\
  & & Complex & 12.5\,/\,78.1\,/\,9.40 & 12.5\,/\,78.1\,/\,9.40 \\
\midrule
C & Kimodo & Common & 46.4\,/\,39.3\,/\,14.3 & 53.6\,/\,28.6\,/\,17.9 \\
  & & Complex & 62.1\,/\,20.7\,/\,17.2 & 58.6\,/\,24.1\,/\,17.2 \\
\bottomrule
\end{tabular}
\end{table}

We evaluate whether video foundation models generate more natural and semantically meaningful human motion than dedicated motion models. We distinguish semantic correctness and action coverage from physical motion quality, since low jerk or foot sliding does not necessarily indicate successful execution of the requested action. We therefore combine VLM-based comparisons, quantitative motion analysis, and human perceptual evaluation. Animated examples are provided in the supplementary video.

\vspace{2pt}
\noindent{\textbf{Experimental Setup.}} We categorize actions as common or complex, including simultaneous and sequential movements (both use $K=5$ keyframes). We compare the video-to-motion pipeline (V2M) against Kimodo and M-GPT3 in A, Kimodo and CondMDI in B, and Kimodo in C. Video-generated motion is reconstructed using GVHMR, and all outputs share the same 3D skeletal representation.

\subsection{Semantic Alignment and Naturalness}
\label{sec:exp_semantic}
We assess semantic alignment (H1, H3) and motion naturalness (H2) through pairwise VLM comparisons using Qwen3-VL-32B-Instruct, following the protocol described in ~\cref{sec:supp_vlm} of the supplementary material. This evaluation contains 65, 64, and 57 prompts in Exp. A, B, and C, respectively, of which 28, 28, and 25 involve multiple actions. \Cref{tab:pairwise_results} reports win/tie/loss (W/T/L) percentages for V2M against each baseline, separately for common and complex actions.

\begin{figure}[t]
\centering
\includegraphics[width=0.95\linewidth]{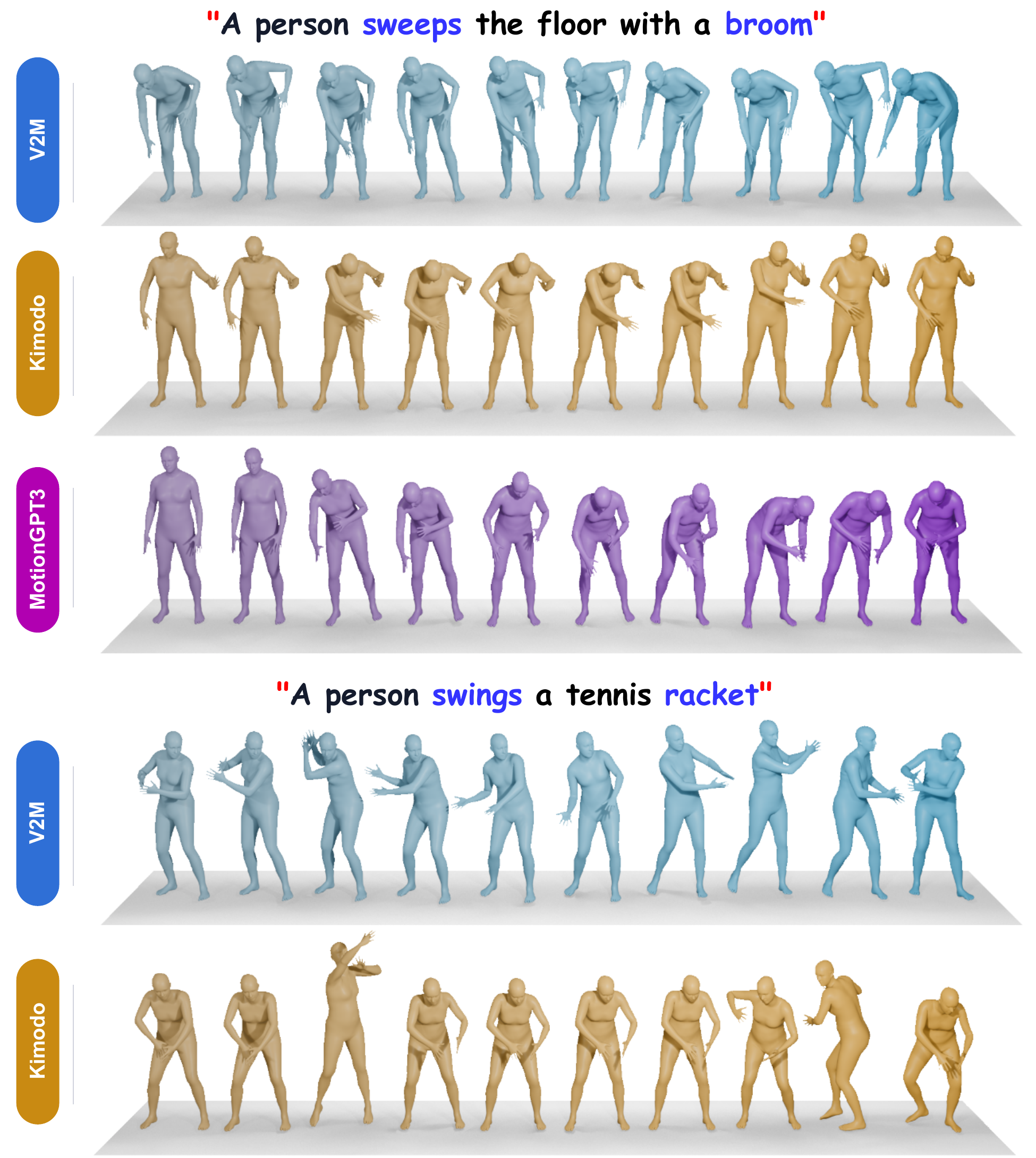}\\[4pt]
\includegraphics[width=0.95\linewidth]{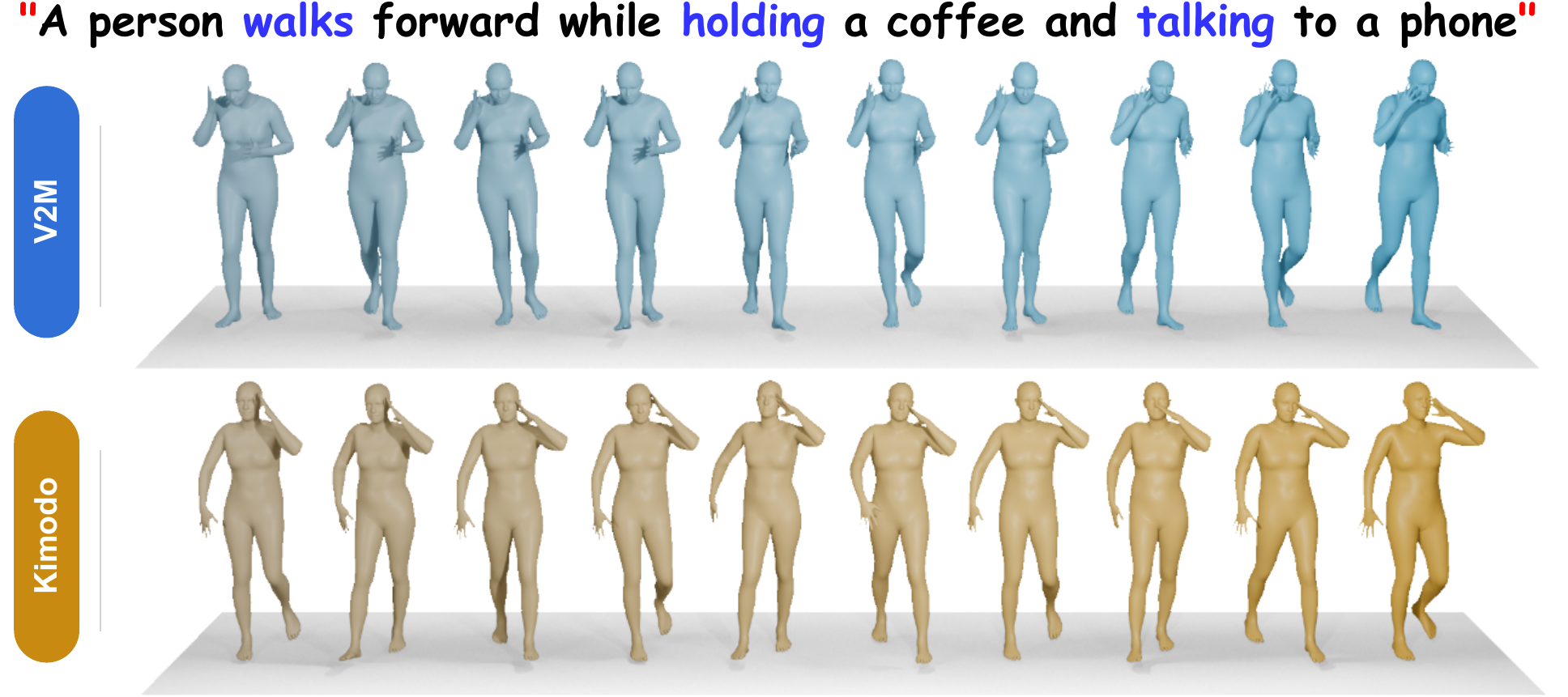}
\caption{\textbf{Experiment A: Prompt-only generation.} Sweeping highlights coordinated arm and body movements, while tennis illustrates racket-swing execution. Walking while talking and holding a coffee evaluates action coverage in compound movements. V2M produces more natural, complete, and semantically coherent motions that better align with the text prompt.}
\label{fig:exp_a}
\vspace{-6pt}
\end{figure}

\vspace{2pt}
\noindent{\textbf{Experiment A: Prompt-Only Generation.}} V2M is preferred over Kimodo for text matching (58.5\% wins) and naturalness (56.9\%). Against M-GPT3, V2M wins 46.2\% and 44.6\%, respectively, with stronger results on complex actions (53.1\% for both criteria, compared with 39.4\% and 36.4\% on common actions). These findings suggest that video models generate more natural and semantically coherent movements, particularly for complex actions (\cref{fig:exp_a}). For instance, V2M successfully combines walking, talking on the phone, and carrying a coffee mug, whereas dedicated motion models struggle to reproduce all actions. Results also suggest a stronger understanding of body-object relationships, such as coordinated hand and foot placement when climbing a ladder or maintaining a consistent grasp without penetrating a carried box (see suppl. video).

\begin{figure}[t]
\centering
\includegraphics[width=\linewidth]{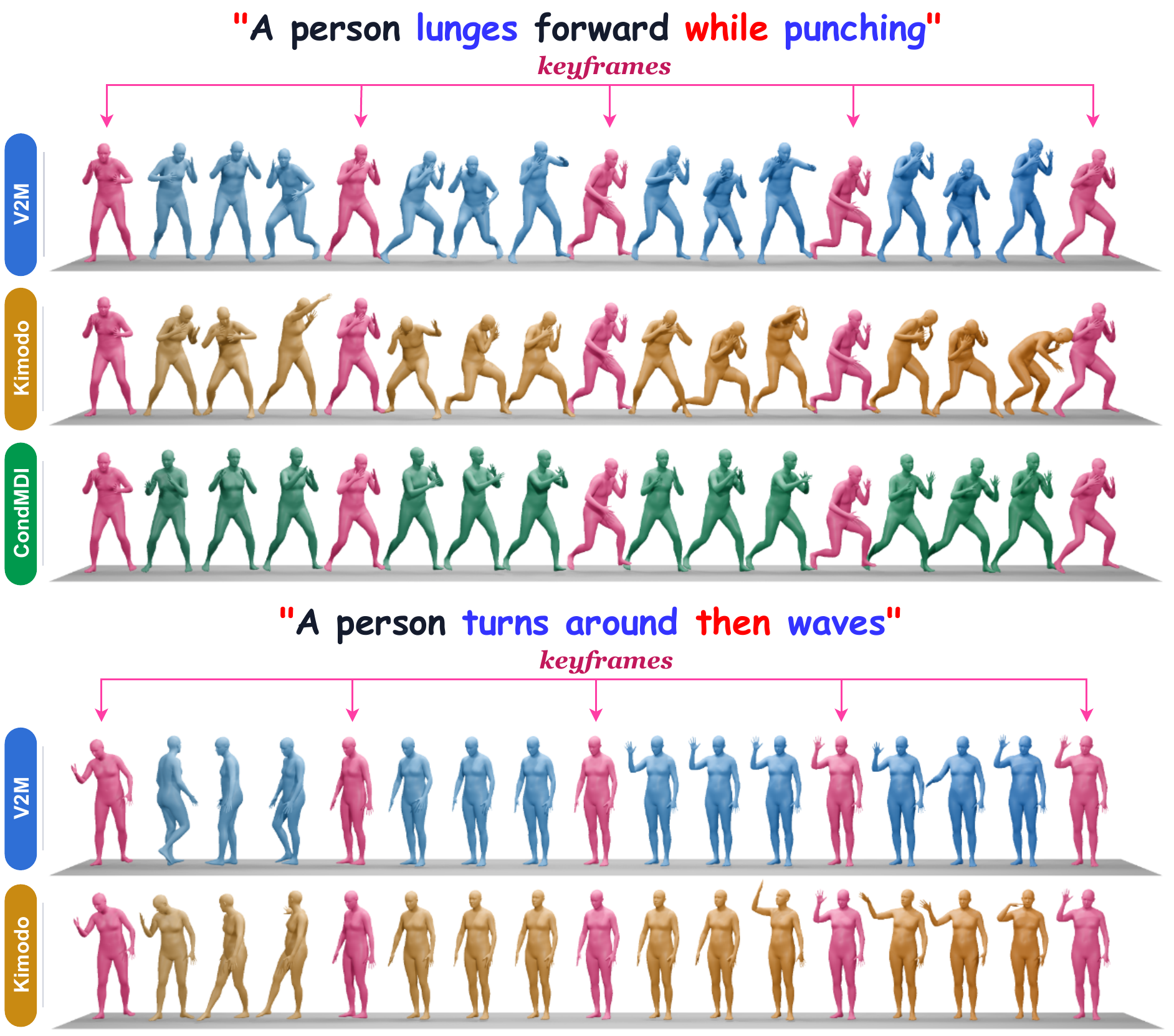}
\caption{\textbf{Experiment B: Video-to-motion.} Lunge-and-punch tests simultaneous movements, while turn-then-wave examines action ordering. Dedicated motion models generate intermediate movements without consistently improving the original video-derived animation, while they may fail to reproduce its complete semantic content.}
\label{fig:exp_b}
\vspace{-6pt}
\end{figure}

\vspace{2pt}
\noindent{\textbf{Experiment B: Video-to-Motion.}} Kimodo and CondMDI reconstruct V2M motion from five sparse keyframes. The comparisons show no consistent improvement over the original V2M motion, with most judgments resulting in ties: 82.8\% and 73.4\% for text matching, and 84.4\% and 70.3\% for naturalness, against Kimodo and CondMDI, respectively. These results suggest that dedicated motion models do not consistently improve the original motion and may occasionally fail to reproduce its complete semantic content, despite being conditioned on its keyframes (\cref{fig:exp_b}). 

\begin{figure}[t]
\centering
\includegraphics[width=\linewidth]{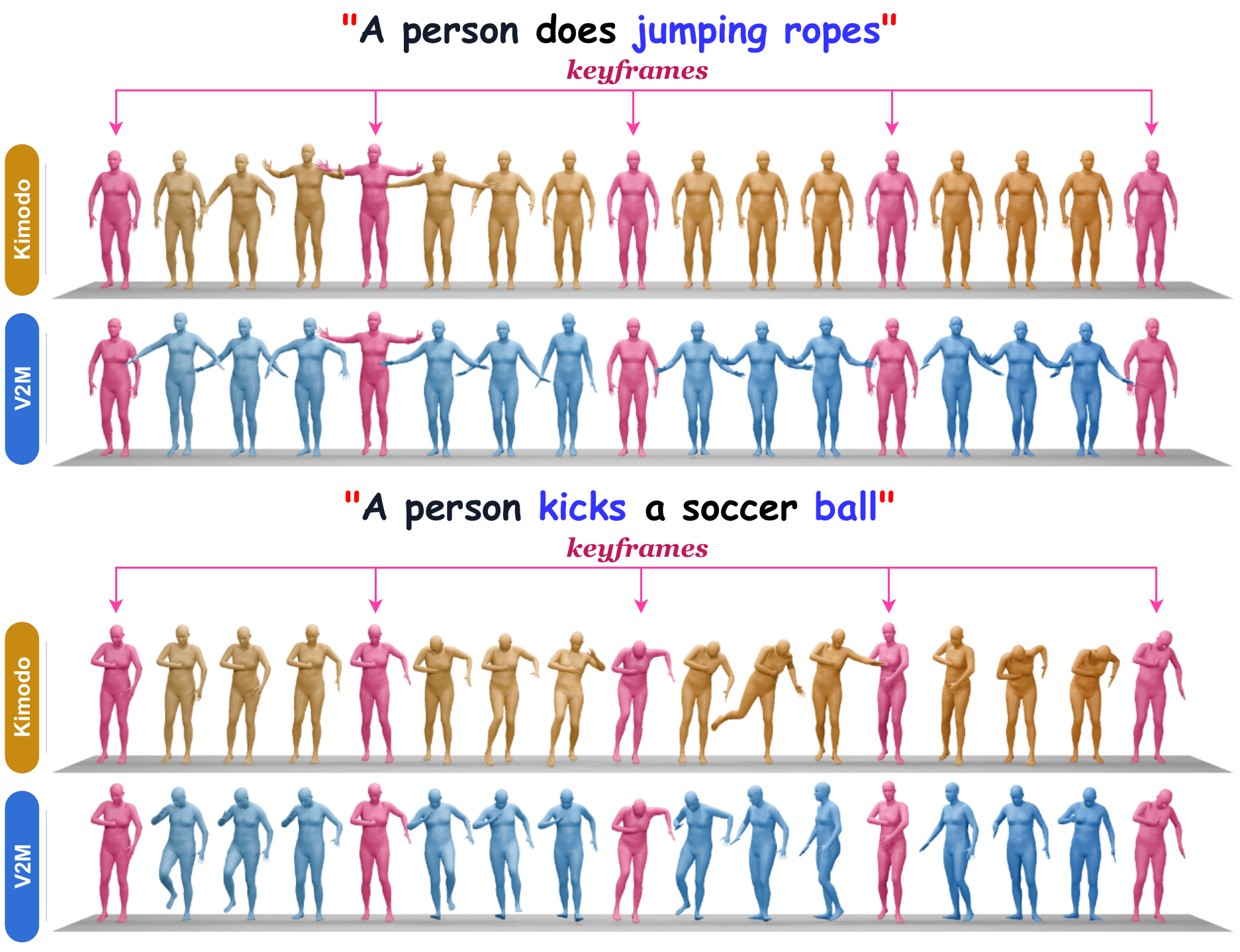}
\caption{\textbf{Experiment C: Motion-to-video.} V2M generates more complete and semantically meaningful movements, both in jumping rope and kicking a soccer ball, illustrating the potential of video models to enrich skeletal animations under loose control.}
\label{fig:exp_c}
\vspace{-4pt}
\end{figure}

\vspace{2pt}
\noindent{\textbf{Experiment C: Motion-to-Video.}} V2M uses Kimodo-generated keyframes as guidance and is preferred over the original Kimodo motion for text matching (54.4\% wins) and naturalness (56.1\%). The advantage increases for complex actions, reaching 62.1\% and 58.6\%, respectively. These findings suggest that video models can enrich skeletal animations with more meaningful and natural movements under loose skeletal control (\cref{fig:exp_c}). 

\subsection{Motion Quality and Fidelity}
\label{sec:exp_physical}

We complement perceptual and semantic evaluation with quantitative motion-quality metrics (H2) and keyframe fidelity measurements (H3).
We measure estimated ground-contact coverage ($Y\leq5$\,cm), positional jerk for selected joints (hands, feet, and pelvis), angular acceleration RMS, and foot sliding. \Cref{tab:motion_metrics} summarizes these results. For B and C, we also evaluate keyframe fidelity using MPJPE, PA-MPJPE, and geodesic rotation errors.

\begin{table}[t]
\centering
\caption{\textbf{Motion quality metrics.} Contact coverage is descriptive. Arrows indicate lower is better, although lower values do not necessarily imply more natural motion. Bold highlights V2M when it achieves the lowest error in each experiment.}
\label{tab:motion_metrics}
\scriptsize
\setlength{\tabcolsep}{2pt}
\renewcommand{\arraystretch}{1.08}
\resizebox{\columnwidth}{!}{%
\begin{tabular}{@{}llrrrrrr@{}}
\toprule
 & & \textbf{Contact} & \multicolumn{3}{c}{\textbf{Positional jerk (m/s$^3$) $\downarrow$}} & \textbf{Ang. RMS} & \textbf{Foot sliding} \\
\cmidrule(lr){4-6}
\textbf{Exp.} & \textbf{Method} & \textbf{(\%)} & \textbf{Hands} & \textbf{Feet} & \textbf{Pelvis} & \textbf{($^\circ$/s$^2$) $\downarrow$} & \textbf{(cm/s) $\downarrow$} \\
\midrule
A & V2M     & 44.15 & 185.20 & 90.55 & 57.23 & \textbf{466.80} & 12.30 \\
  & Kimodo  & 72.77 & \textbf{95.93} & \textbf{80.90} & 40.87 & 1332.00 & \textbf{6.19} \\
  & M-GPT3  & 64.07 & 108.90 & 96.50 & \textbf{36.78} & 2608.00 & 12.56 \\
\midrule
B & V2M     & 34.36 & 231.70 & \textbf{82.60} & \textbf{70.24} & \textbf{770.70} & \textbf{7.47} \\
  & Kimodo  & 64.66 & \textbf{151.30} & 126.20 & 78.60 & 1724.00 & 7.82 \\
  & CondMDI & 55.78 & 249.50 & 205.30 & 185.80 & 1716.00 & 17.07 \\
\midrule
C & V2M     & 41.06 & 119.40 & \textbf{80.54} & 70.60 & \textbf{351.40} & \textbf{7.91} \\
  & Kimodo  & 85.68 & \textbf{81.32} & 97.79 & \textbf{35.23} & 982.50 & 11.92 \\
\bottomrule
\end{tabular}%
}
\end{table}

\vspace{2pt}
\noindent{\textbf{Ground-Contact Coverage.}} 
V2M exhibits lower estimated ground-contact coverage than dedicated motion models (44.15\%, 34.36\%, and 41.06\% in A, B, and C). This may reflect more dynamic movements and frequent foot lifting, while dedicated models sometimes produce limited or idle motion (see supplementary video). Note, coverage does not directly measure action completion.

\vspace{2pt}
\noindent{\textbf{Motion Smoothness and Foot Sliding.}} V2M generally exhibits higher positional jerk but lower angular acceleration RMS than Kimodo, while foot sliding is higher in A but lower in B and C. These mixed results highlight the limitations of kinematic metrics: dedicated motion models may achieve low errors through limited or idle movements rather than successfully executing the requested actions. Consequently, these metrics should be interpreted alongside semantic alignment and perceptual naturalness. Indeed, our human study (\cref{sec:exp_userstudy}) consistently favors V2M for motion naturalness. Additional examples are provided in the supplementary video.

\vspace{2pt}
\noindent{\textbf{Keyframe Fidelity.}} In B, V2M serves as the reference motion from which keyframes are extracted. Kimodo and CondMDI achieve MPJPE of 3.04\,cm and 12.17\,cm, respectively, with 22-joint geodesic rotation errors of 15.39$^\circ$ and 10.75$^\circ$, and root rotation errors of 2.25$^\circ$ and 4.03$^\circ$. These results indicate relatively small positional deviations from the reference keyframes. In C, V2M is guided by 2D projections of Kimodo keyframes. The reconstructed 3D motion achieves MPJPE of 15.03\,cm, PA-MPJPE of 7.10\,cm, joint rotation error of 20.57$^\circ$, and root rotation error of 12.60$^\circ$. These errors may partly reflect the depth ambiguity of 2D guidance and the additional 3D reconstruction step. Importantly, lower keyframe errors do not necessarily imply better motion under loose control, where deviations may enrich intermediate movements while retaining the intended action.

\subsection{Human Perceptual Evaluation}
\label{sec:exp_userstudy}

We conduct an online perceptual study comparing V2M against Kimodo across 9, 8, and 9 animation pairs from Exp. A, B, and C, respectively. A total of 43 participants completed the study. Participants assessed motion naturalness (H2), semantic alignment and action coverage (H1, H3), and visible motion artifacts (H2), with the detailed protocol provided in ~\cref{sec:supp_human} of the supplementary material. We report win/tie/loss (W/T/L) percentages from the perspective of V2M for pairwise comparisons, and individual action-completion rates for each model.

\Cref{tab:perceptual_study} summarizes the results. V2M is preferred over Kimodo for motion naturalness in 72\%, 75\%, and 61\% of comparisons in A, B, and C, respectively, and for semantic alignment in 88\%, 74\%, and 66\%. Participants also favor V2M for fewer visible artifacts (63\%, 72\%, and 54\%). These preferences support H2, indicating that video-generated movements are perceived as more natural, semantically meaningful, and visually coherent, even when quantitative motion-quality metrics show mixed results. While the VLM evaluation reports predominantly tied comparisons, the human study reveals a clearer preference for V2M, suggesting differences in the sensitivity of the two evaluation protocols.

\begin{table}[t]
\centering
\caption{\textbf{Human perceptual evaluation.} Pairwise W/T/L (\%) for V2M against Kimodo, and action-completion rates (\%) for video (V) and motion (M) models.}
\label{tab:perceptual_study}
\footnotesize
\setlength{\tabcolsep}{3pt}
\renewcommand{\arraystretch}{1.10}
\begin{tabular}{@{}lcccc@{}}
\toprule
\textbf{Exp.} & \textbf{Naturalness} & \textbf{Semantic} &
\textbf{Fewer artifacts} & \textbf{Action completion} \\
 & \textbf{W/T/L} & \textbf{W/T/L} &
\textbf{W/T/L} & \textbf{V/M (\%)} \\
\midrule
A & 72 / 15 / 13 & 88 / 7 / 5 & 63 / 25 / 12 & 86 / 38 \\
B & 75 / 14 / 11 & 74 / 10 / 16 & 72 / 17 / 11 & 84 / 63 \\
C & 61 / 20 / 19 & 66 / 16 / 18 & 54 / 31 / 15 & 86 / 63 \\
\bottomrule
\end{tabular}%
\end{table}

\vspace{2pt}

\noindent{\textbf{Action Coverage.}} Beyond semantic alignment, we assess whether animations execute all requested actions, particularly for simultaneous and/or sequential movements, where dedicated motion models may prioritize the dominant movement while omitting others. V2M achieves action-completion scores of 86\%, 84\%, and 86\% in Exp. A, B, and C, respectively, compared with 38\%, 63\%, and 63\% for Kimodo. For compound actions, V2M also better preserves temporal ordering and simultaneity (86\%, 86\%, and 84\% versus 45\%, 69\%, and 66\%). These findings support H1, demonstrating broader semantic motion capabilities, while the advantages under sparse guidance in B and C provide additional evidence for H3.

\subsection{Discussion}
\label{sec:exp_discussion}

Our results suggest that video foundation models possess richer motion priors than dedicated motion models, particularly for complex and compound actions. In text-only generation (A), V2M achieves better semantic alignment, perceived naturalness, and action coverage, demonstrating its ability to execute multiple requested movements (H1). Although quantitative metrics show mixed results, perceptual evaluations favor video-generated motion in terms of naturalness and fewer visible artifacts (H2). Under sparse guidance, dedicated motion models do not improve V2M motion (B), whereas video models enrich motion-model animations with more natural and semantically meaningful movements (C). Keyframe fidelity measures how well sparse guidance is preserved, without necessarily reflecting motion quality (H3). These findings suggest complementary roles: video foundation models provide rich semantic and motion priors for generation, while dedicated motion models offer capabilities for refinement and structural control. The supplementary video illustrates these differences through representative examples.

\section{Conclusions and Future Challenges}
\label{sec:conclusion}
This work advances the view that video foundation models have become powerful motion generators, with richer and more general motion priors than dedicated motion models. Our experiments suggest that their advantage extends beyond visual quality to motion naturalness, semantic coherence, and the ability to perform complex and compound actions. While dedicated motion models do not consistently improve video-generated motion, video models can enrich skeletal animations under sparse guidance. \emph{Rather than relying exclusively on dedicated motion models, video foundation models can leverage richer semantic and motion priors for generation, while dedicated motion models complement them through refinement and control.}

This perspective raises several challenges. First, the motion knowledge embedded in video foundation models remains implicit, and extracting accurate, editable 3D animations remains difficult, despite continuous advances in motion reconstruction methods. These limitations are particularly evident in fast movements, occlusions, and complex interactions, as also noted in previous work~\cite{janson2024towards,chefer2025videojam}. Second, exploiting these rich motion priors requires new forms of control that preserve the intended action without overconstraining the generated movement. Loose skeletal guidance~\cite{loosecontrol,loosecontrolvideo} offers an initial direction, but more expressive and adaptive control mechanisms are needed. Finally, the broader challenge is to extend this perspective beyond individual human motion, toward the generation and control of interacting characters, objects, and environments. \emph{Ultimately, we envision a future in which animation is no longer primarily about generating motion from scratch, but about directing and shaping the rich motion knowledge already embedded in video foundation models.}

{
    \small
    \bibliographystyle{ieeenat_fullname}
    \bibliography{mybibliography}
}

\clearpage
\maketitlesupplementary

\setcounter{section}{0}
\renewcommand{\thesection}{S\arabic{section}}
\renewcommand{\thesubsection}{\thesection.\arabic{subsection}}
\renewcommand{\thesubsubsection}{\thesubsection.\arabic{subsubsection}}

\providecommand{\theHsection}{}
\providecommand{\theHsubsection}{}
\providecommand{\theHsubsubsection}{}
\renewcommand{\theHsection}{supp.\arabic{section}}
\renewcommand{\theHsubsection}{supp.\arabic{section}.\arabic{subsection}}
\renewcommand{\theHsubsubsection}{supp.\arabic{section}.\arabic{subsection}.\arabic{subsubsection}}

\setcounter{figure}{0}
\setcounter{table}{0}
\setcounter{equation}{0}
\renewcommand{\thefigure}{S\arabic{figure}}
\renewcommand{\thetable}{S\arabic{table}}
\renewcommand{\theequation}{S\arabic{equation}}
\providecommand{\theHfigure}{}
\providecommand{\theHtable}{}
\providecommand{\theHequation}{}
\renewcommand{\theHfigure}{supp.\arabic{figure}}
\renewcommand{\theHtable}{supp.\arabic{table}}
\renewcommand{\theHequation}{supp.\arabic{equation}}
\section{Evaluation Metrics and Implementation Details}
\label{sec:supp_metrics}

We evaluate generated motion through semantic alignment, perceptual naturalness, physical motion quality, and keyframe fidelity, as described in \cref{sec:evaluation} of the main paper. All motions are represented using the same 22-joint SMPL skeleton at 30\,fps, with a $Y$-up coordinate system and the ground plane at $Y=0$. We denote the 3D position of joint $j$ at frame $t$ by $\mathbf{p}_{t,j}$ and the frame interval by $\Delta t=1/30$\,s. Unless otherwise specified, metrics are computed per sequence and then averaged across the evaluated sequences with equal weight.

\subsection{Semantic Alignment and Motion Naturalness}
\label{sec:supp_vlm}

\noindent\textbf{VLM-based Pairwise Evaluation.}
We use Qwen3-VL-32B-Instruct as an automated judge to compare the semantic alignment and perceived naturalness of motions generated by V2M and dedicated motion models. The evaluation includes 65, 64, and 57 text prompts in Experiments A, B, and C, respectively. Each comparison uses two animations generated for the same prompt, represented in a common 3D skeletal format and rendered under identical conditions.

\noindent\textbf{Motion Rendering and Blinding.}
All motions are represented using the neutral SMPL body ($\boldsymbol{\beta}=\mathbf{0}$) in a gravity-aligned, $Y$-up coordinate system at 30~fps. 
Animations are rendered as uniformly colored grey characters on a flat floor, using fixed lighting and a static camera. For each prompt, the camera is fitted jointly to the motions produced by all compared methods, ensuring the same viewpoint and framing across animations. Objects referenced in the text, such as balls or tools, are not rendered; their associated actions are evaluated from the character's body movements. All clips are assigned anonymous identifiers, and model identities and source filenames are withheld from the judge.

\noindent\textbf{Pairwise Judging.}
For each comparison, we provide the VLM with the original text prompt and sampled frames from each animation, resized to $576\times576$ pixels. The judge first describes the observed movements in each clip, then independently answers two questions: (i) which animation better follows the text, considering all requested actions, their temporal order, and any simultaneous movements; and (ii) which animation exhibits more natural human motion, considering foot sliding, floating, jitter, implausible poses, and stiff body parts. The same instructions are used for all model pairs. Inference uses deterministic greedy decoding.

\noindent\textbf{Preference Aggregation.}
To reduce presentation-order bias, each pair is evaluated twice, with the animation order reversed. A win or loss is recorded only when both evaluations agree on the same preferred method; explicit ties and disagreements between orders are counted as ties. We report win/tie/loss (W/T/L) percentages from the perspective of V2M. Results are further separated into \emph{common} and \emph{complex} prompts: the latter comprise actions explicitly categorized as complex and prompts requiring sequential or simultaneous movements, while the former comprise the remaining single-action prompts. We additionally compute net preference (wins minus losses) and its 95\% confidence interval using 10,000 prompt-level bootstrap resamples. These automated judgments evaluate perceived semantic and motion quality, and are interpreted alongside the human perceptual study and quantitative motion metrics.

\noindent\textbf{VLM Judge Prompt.}
For reproducibility, we provide the pairwise evaluation
prompt used by the VLM. The same instructions are
applied across all experiments and model comparisons,
with only the text description and input animations
changed. The judge first describes each motion
independently before comparing semantic alignment
and perceived naturalness. Below, we show the prompt
instantiated for a representative sequential action.

\begin{tcolorbox}[
    colback=black!5,
    colframe=black!15,
    boxrule=0.4pt,
    arc=1mm,
    left=6pt,
    right=6pt,
    top=6pt,
    bottom=6pt,
    breakable
]
\footnotesize
\raggedright

You are an expert in human motion. Each video shows
a 3D mannequin moving on a flat floor, filmed by a
fixed camera; both videos use the same camera, body
and lighting. Nothing else is rendered: objects,
tools, balls, stairs, ladders, walls or bars never
appear, even when the person interacts with them.
At the start the person faces roughly toward the
camera, so the person's own LEFT side appears on
the RIGHT of the image; moving forward means
moving toward the camera (the figure grows larger
and its feet move down in the image).

\medskip
The two videos were generated for the same text:
``A person marches in place, then salutes.''.

Watch both videos completely.

\medskip
\textbf{Step 1.}
Describe what the person does in Video 1, in order,
in 2-3 sentences.

\smallskip
\textbf{Step 2.}
Describe what the person does in Video 2, in order,
in 2-3 sentences.

\smallskip
\textbf{Step 3.}
\texttt{text\_match}: which video performs the text
better? Check every part of the text (each action,
the order of actions, actions done at the same time,
and the body movements implied by objects, which
are not rendered).

\smallskip
\textbf{Step 4.}
\texttt{natural}: which video looks more like a real
person moving? Look for sliding feet, floating,
jitter, impossible poses, and stiff or frozen
body parts.

\medskip
Answer ``tie'' only if the videos are truly equal.

\medskip
\textbf{Return only JSON:}

\smallskip
{\ttfamily\scriptsize
\{"video1": "...", "video2": "...",\par
"text\_match": "1|2|tie",\par
"natural": "1|2|tie",\par
"reason": "one sentence"\}
}

\end{tcolorbox}

\subsection{Physical Motion Quality}
\label{sec:supp_physical}

We evaluate motion quality through positional jerk, rotational smoothness, local jitter, ground-contact coverage, foot sliding, and floor penetration. These metrics characterize motion dynamics and potential artifacts but do not independently establish perceptual naturalness or action correctness. In particular, low kinematic errors may also result from limited or idle movements.

\paragraph{Positional Jerk (m/s$^3$) $\downarrow$.}
We measure abrupt changes in joint acceleration using the third temporal derivative of joint positions, approximated by finite differences:
\begin{equation}
\mathbf{j}_{t,j} =
\frac{
\mathbf{p}_{t+3,j}
-3\mathbf{p}_{t+2,j}
+3\mathbf{p}_{t+1,j}
-\mathbf{p}_{t,j}
}{\Delta t^3}.
\end{equation}
The average jerk magnitude over a set of joints $\mathcal{J}$ is:
\begin{equation}
J_{\mathcal{J}} =
\frac{1}{|\mathcal{T}|\,|\mathcal{J}|}
\sum_{t\in\mathcal{T}}
\sum_{j\in\mathcal{J}}
\|\mathbf{j}_{t,j}\|_2,
\end{equation}
where $\mathcal{T}$ contains valid frames. We report average jerk for six joints (left and right wrists, left and right ankles, head, and pelvis), together with separate averages for the hands and feet and individual values for the head and pelvis. Root-relative jerk is additionally computed by subtracting the root position from each joint trajectory before differentiation. Higher jerk may indicate abrupt motion artifacts but may also reflect more dynamic movements.

\paragraph{Angular Acceleration RMS ($^\circ$/s$^2$) $\downarrow$.}
We evaluate rotational smoothness using joint rotations $\mathbf{R}_{t,j}\in SO(3)$. Angular velocity is approximated from consecutive rotations using the logarithmic map:
\begin{equation}
\boldsymbol{\omega}_{t,j} =
\frac{
\operatorname{Log}(
\mathbf{R}_{t,j}^{\top}\mathbf{R}_{t+1,j}
)
}{\Delta t},
\end{equation}
where $\operatorname{Log}$ returns a rotation vector. Angular acceleration is approximated as:
\begin{equation}
\boldsymbol{\alpha}_{t,j} =
\frac{
\boldsymbol{\omega}_{t+1,j}
-\boldsymbol{\omega}_{t,j}
}{\Delta t}.
\end{equation}
We compute the root-mean-square angular acceleration across five joints (left and right wrists, left and right ankles, and pelvis), excluding the head:
\begin{equation}
A_{\mathrm{RMS}} =
\sqrt{
\frac{1}{|\mathcal{T}|\,|\mathcal{J}_5|}
\sum_{t\in\mathcal{T}}
\sum_{j\in\mathcal{J}_5}
\|\boldsymbol{\alpha}_{t,j}\|_2^2
}.
\end{equation}
The result is converted to degrees per second squared. We additionally evaluate rotational smoothness across all 22 joints and individual joints.

\paragraph{Local Motion Jitter (cm) $\downarrow$.}
We measure high-frequency variations in root-relative joint trajectories using a seven-frame, second-order Savitzky--Golay filter (SG7). Let $\mathbf{q}_{t,j}$ denote the root-relative joint position and $\tilde{\mathbf{q}}_{t,j}$ its smoothed trajectory. Jitter is computed as the root-mean-square residual:
\begin{equation}
\mathrm{Jitter} =
100\sqrt{
\frac{1}{|\mathcal{T}|\,|\mathcal{J}_6|}
\sum_{t\in\mathcal{T}}
\sum_{j\in\mathcal{J}_6}
\|\mathbf{q}_{t,j}-\tilde{\mathbf{q}}_{t,j}\|_2^2
}.
\end{equation}
We evaluate six joints (wrists, ankles, head, and pelvis), reporting the result in centimeters. This metric captures local irregularities but may also penalize rapid, intentional movements.

\paragraph{Ground-Contact Coverage (\%).}
We estimate potential foot-ground contact using both vertical foot position and velocity. A foot is considered in contact when its height is at most 5\,cm and its absolute vertical velocity does not exceed 20\,cm/s:
\begin{equation}
c_{t,f} =
\mathbb{I}\left(
p^Y_{t,f}\leq 0.05
\;\land\;
|v^Y_{t,f}|\leq 0.20
\right),
\end{equation}
where $f$ denotes the left or right foot, position is measured in meters, and velocity in m/s. We report the percentage of foot-frame observations satisfying these conditions:
\begin{equation}
\mathrm{GCC} =
\frac{100}{2T}
\sum_{t=1}^{T}
\sum_{f\in\{L,R\}}c_{t,f}.
\end{equation}
Contact coverage is descriptive rather than an error measure. Lower coverage may indicate more dynamic or airborne movements, but can also reflect inaccurate foot placement or reconstruction artifacts. We additionally evaluate height thresholds of 7.5, 10, 12.5, and 15\,cm.

\paragraph{Foot Sliding (cm/s) $\downarrow$.}
We measure horizontal foot velocity during estimated ground contact. For each foot, horizontal velocity is:
\begin{equation}
v^{XZ}_{t,f} =
\left\|
\frac{
\mathbf{p}^{XZ}_{t+1,f}
-\mathbf{p}^{XZ}_{t-1,f}
}{2\Delta t}
\right\|_2,
\end{equation}
Let $\mathcal{C}$ contain valid foot-frame observations satisfying both the height ($Y\leq0.05$\,m) and vertical velocity ($|v^Y|\leq0.20$\,m/s) conditions. Foot sliding is calculated as:
\begin{equation}
\mathrm{FS} =
\frac{100}{|\mathcal{C}|}
\sum_{(t,f)\in\mathcal{C}}
v^{XZ}_{t,f},
\end{equation}
where the factor 100 converts m/s to cm/s. Sequences without valid contact samples are excluded from the aggregate sliding mean. Lower values generally indicate more stable foot placement, although the metric should be interpreted alongside estimated contact coverage. We additionally report results for alternative contact-height thresholds.

\paragraph{Floor Penetration (cm) $\downarrow$.}
We estimate foot-joint penetration below the ground plane ($Y<0$), defining the penetration depth as:
\begin{equation}
d_{t,f} =
100\max(0,-p^Y_{t,f}).
\end{equation}
We report the maximum penetration depth, mean depth among penetrating samples, mean depth across all foot samples, and the numbers of penetration episodes, frames, and foot-joint samples. A penetration episode corresponds to a continuous interval during which one foot joint remains below the ground plane. Since these measurements use joint positions rather than the complete foot mesh, they provide approximate indicators of ground-intersection artifacts.

\subsection{Keyframe Fidelity}
\label{sec:supp_fidelity}

We evaluate keyframe fidelity in Experiments B and C, which employ different conditioning representations. In Experiment B, dedicated motion models receive full 3D skeletal keyframes extracted from V2M motion. In Experiment C, the video model receives their 2D projections from Kimodo-generated motion, and the resulting video is subsequently reconstructed into 3D. We compare the generated joint positions and rotations against the corresponding original 3D reference poses.

Let $\mathcal{K}$ denote the set of temporally matched reference keyframes, with generated positions $\hat{\mathbf{p}}_{t,j}$ and reference positions $\mathbf{p}_{t,j}$.

\paragraph{Mean Per-Joint Position Error (MPJPE) $\downarrow$.}
We measure the average Euclidean distance between corresponding generated and reference joints:
\begin{equation}
\mathrm{MPJPE} =
\frac{100}{|\mathcal{K}|J}
\sum_{t\in\mathcal{K}}\sum_{j=1}^{J}
\|\hat{\mathbf{p}}_{t,j}-\mathbf{p}_{t,j}\|_2.
\end{equation}
where $J=22$. MPJPE is reported in centimeters.

\paragraph{Procrustes-Aligned MPJPE (PA-MPJPE) $\downarrow$.}
We additionally compute MPJPE after independently aligning each generated keyframe to its corresponding reference pose using an optimal similarity transformation, including rotation, translation, and uniform scaling. This reduces global alignment discrepancies and measures the residual joint-position differences. PA-MPJPE is reported in centimeters.

\paragraph{Joint Rotation Error ($^\circ$) $\downarrow$.}
We evaluate rotational fidelity using the geodesic distance between generated and reference rotations:
\begin{equation}
d_R(\hat{\mathbf{R}},\mathbf{R}) =
\cos^{-1}\left(
\operatorname{clip}\left(
\frac{
\operatorname{tr}(\hat{\mathbf{R}}^\top\mathbf{R})-1
}{2},
-1,1
\right)\right).
\end{equation}
We report the mean geodesic error across all 22 joints and matched keyframes, converted to degrees. We additionally report the mean error across the 21 non-root joints and the root orientation error separately.

\paragraph{Temporal Matching and Interpretation.}
Only reference keyframes with valid corresponding frames in the generated sequence are evaluated; unmatched keyframes are excluded. Metrics are computed per sequence and then averaged with equal weight, with the numbers of matched and reference keyframes recorded separately.

Importantly, Experiments B and C do not impose equivalent constraints. Experiment B directly conditions skeletal motion on 3D poses, whereas Experiment C uses sparse 2D projections that do not uniquely specify depth or the complete 3D joint configuration. Furthermore, Experiment C introduces an additional video-to-3D reconstruction stage. Consequently, differences in fidelity may partly reflect projection ambiguity and reconstruction errors. These metrics measure structural consistency at selected keyframes rather than naturalness or preservation of the complete motion trajectory.

\subsection{Human Perceptual Evaluation}
\label{sec:supp_human}

\noindent\textbf{Study Setup.}
We conduct a blinded online perceptual study comparing V2M with Kimodo using pairs of 3D skeletal animations. Each pair is presented alongside the text description used to generate the motions. Both animations are rendered with neutral-colored avatars under identical visualization conditions, without identifying the underlying models. The video- and motion-model animations appear on both sides across the comparison set rather than consistently occupying the same position. A total of 43 participants completed the study, each evaluating 26 animation pairs. The selection and presentation order of the pairs vary across participants.

\noindent\textbf{Pairwise Questions.}
For each pair, participants answer three questions: (i) ``Which motion better matches the description?'', (ii) ``Which motion looks more like a real person moving?'', and (iii) ``Which motion has fewer visible errors?'' These questions assess semantic alignment, perceived naturalness, and visible motion artifacts, respectively. The semantic question concerns how well the observed motion reflects the requested action, while the other two distinguish the overall impression of human movement from more localized errors, such as foot sliding, jitter, or implausible body poses. The questions are kept the same across all three experiments.

\noindent\textbf{Action Coverage.}
To examine whether an animation performs the requested movements rather than only the dominant action, we also use a prompt-specific checklist for actions that can be evaluated separately. Participants assess the individual components of the description for each animation, without directly comparing the two models. For instance, a prompt asking a person to turn around and then wave is evaluated in terms of both movements. For compound prompts, an additional item assesses whether the movements occur in the required order or at the same time. This allows us to distinguish the presence of individual actions from their temporal coordination.

\noindent\textbf{Preference Reporting and Scoring.}
For each pairwise question, participants select ``Left'', ``Right'', ``About the same'', or ``Cannot judge''. A preference for V2M is counted as a win and a preference for Kimodo as a loss. In the win/tie/loss (W/T/L) results reported in the main paper, ``About the same'' and ``Cannot judge'' are grouped in the non-preference category, reported as ties. Thus, the percentages are computed over all completed responses for each question; the displayed tie percentage is obtained after rounding the win and loss percentages so each row sums to 100\%.

For the action checklist, participants indicate ``Yes'', ``Partly'', ``No'', or ``Cannot judge'' for each component. We assign scores of $1$, $0.5$, and $0$ to ``Yes'', ``Partly'', and ``No'', respectively, and exclude ``Cannot judge'' from the corresponding average. We compute the action-completion score separately for video- and motion-model animations by averaging valid individual-action scores and reporting them as percentages. Temporal ordering and simultaneity are scored in the same way but reported separately from action completion. Incomplete submissions are excluded. Participants are not shown the original source motion as a reference; preservation of skeletal guidance is evaluated separately through quantitative keyframe fidelity.

\end{document}